# DeepGRU: Deep Gesture Recognition Utility

Mehran Maghoumi[1,2], Joseph J. LaViola Jr.[2]

[1]NVIDIA

[2]University of Central Florida

{mehran,jjl}@cs.ucf.edu

## Abstract

*We propose DeepGRU, a novel end-to-end deep network model informed by recent developments in deep learning for gesture and action recognition, that is streamlined and device-agnostic. DeepGRU, which uses only raw skeleton, pose or vector data is quickly understood, implemented, and trained, and yet achieves state-of-the-art results on challenging datasets. At the heart of our method lies a set of stacked gated recurrent units (GRU), two fully-connected layers and a novel global attention model. We evaluate our method on seven publicly available datasets, containing various number of samples and spanning over a broad range of interactions (full-body, multi-actor, hand gestures, etc.). In all but one case we outperform the state-of-the-art pose-based methods. For instance, we achieve a recognition accuracy of 84.9% and 92.3% on cross-subject and cross-view tests of the NTU RGB+D dataset respectively, and also 100% recognition accuracy on the UT-Kinect dataset. While DeepGRU works well on large datasets with many training samples, we show that even in the absence of a large number of training data, and with as little as four samples per class, DeepGRU can beat traditional methods specifically designed for small training sets. Lastly, we demonstrate that even without powerful hardware, and using only the CPU, our method can still be trained in under 10 minutes on small-scale datasets, making it an enticing choice for rapid application prototyping and development.*

## 1. Introduction

With the advent of various input devices, gesture recognition has become increasingly relevant in human-computer interaction. As these input devices get more capable and precise, the complexity of the interactions that they can capture also increases which, in turn, ignites the need for recognition methods that can leverage these capabilities. From a practitioners point of view, a gesture recognizer would need to possess a set of traits in order to gain adoption: it should capture the fine differences among gestures and distinguish one gesture from another with a high degree of confidence, while being able to work with a vast number of input devices and gesture modalities. Concurrently, a recognition method should enable system designers to integrate the method into their workflow with the least amount of effort. These goals are often at odds: the recognition power of a recognizer usually comes at the cost of increased complexity and decreased flexibility of working across different input devices and modalities.

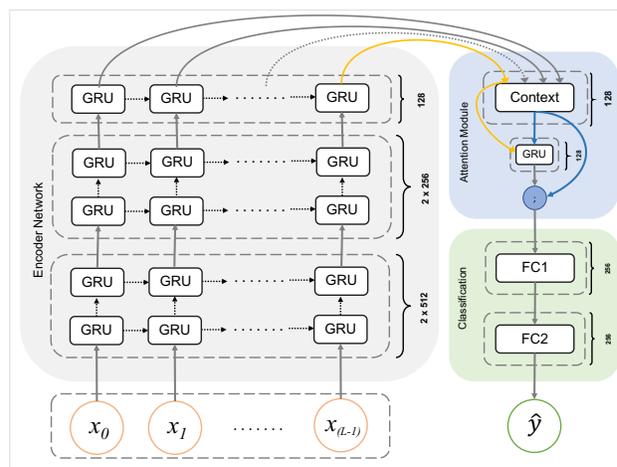

Figure 1. **DeepGRU** – the proposed recurrent model for gesture recognition which consists of an *encoder network* of stacked gated recurrent units (GRU), the *attention module* and the *classification* layers. The input $\mathbf{x} = (x_0, x_1, ..., x_{(L-1)})$ is a sequence of vector data of arbitrary length and the output is the predicted class label $\hat{y}$. The number of the hidden units for each layer is displayed next to every component (see Section 3 for a thorough description).

With these contradicting goals in mind, we introduce *DeepGRU*: an end-to-end deep network-based gesture recognition utility (see Figure 1). DeepGRU works directly with raw 3D skeleton, pose or other vector features (*e.g.* acceleration, angular velocity, *etc.*) produced by noisy commodity hardware, thus requiring minimal domain-specific knowledge to use. With roughly 4 million trainable parameters, DeepGRU is a rather small network by modern standards and is budget-aware when computational power is constrained. Through evaluations on different datasets and gesture modalities, we demonstrate that our proposed method achieves state-of-the-art recognition accuracy on small and large training data alike. We demonstrate the relevance of our deep network model for small-scale problems with limited amount of training data. Specifically, we show that with as little as four training samples per class, our method can produce state-of-the-art results in such settings, and that it is possible to train our model in a reasonable amount of time using only the CPU.

**Contributions.** Our main contributions are devising a novel network model that works with raw vector data and is: **(1)** intuitive to understand and easy to implement, **(2)** easy to use, works out-of-the-box on noisy data, and is easy to train, without requiring powerful hardware **(3)** achieves state-of-the-art results in various use-cases, even with limited amount of training data. We believe **(1)** and **(2)** make DeepGRU enticing for application developers while **(3)** appeals to seasoned practitioners. To our knowledge, no prior work specifically focuses on model simplicity, accessibility for the masses, small training sets or CPU-only training which we think makes DeepGRU unique among its peers.

## 2. Related Work

**Recognition with hand-crafted features.** Despite the success of end-to-end methods, classical methods that use hand-crafted features to perform recognition have been used with great success [18][23][26][27][49][60]. As Cheema *et al.* [11] showed, these methods can achieve excellent recognition results. They compared the performance of five algorithms (AdaBoost, SVM, Bayes, decision trees and the linear classifier) on Wii controller gestures and concluded that, in some cases, the seemingly simple linear classifier can recognize a set of 25 gestures with 99% accuracy. Weng *et al.* [67] leveraged the spatio-temporal relations in action sequences with naïve-Bayes nearest-neighbor classifiers [8] to recognize actions. Xia *et al.* [69] used hidden Markov models (HMM) and the histogram of 3D joint locations to recognize gestures. Vemulapalli *et al.* [59] represented skeletal gestures as curves in a Lie group and used a combination of classifiers to recognize the gestures. Wang *et al.* [65] modeled the spatio-temporal motion properties of joints with a graph of motionlets. These graphs were then classified using SVMs to recognize actions. Evangelidis *et al.* [21] proposed skeletal quads, a skeleton descriptor which encodes relative position of joint quadruples which were then used for classifying actions.

The distinguishing characteristic of our approach compared to all of these methods is that we use the raw data of noisy input devices and do not hand-craft any features. Rather, our encoder network (Section 3.2) learns suitable feature representations during end-to-end training.

**Recurrent architectures.** The literature contains a large body of work that use recurrent neural networks (RNN) for action and gesture recognition [12] [16] [19] [29] [30] [31] [36] [53] [58] [66]. Here, we focus on the ones that are the most closely related to our work.

Shahroudy *et al.* [46] showed the power of recurrent architectures and long-short term memory (LSTM) units [25] for large-scale gesture recognition. Zhang *et al.* [71] proposed a view-adaptive scheme to achieve view-invariant action recognition. Their model consisted of LSTM units that would learn the most suitable transformation of samples to achieve consistent viewpoints. Liu *et al.*[37] incorporated the spatio-temporal and contextual dependencies to recognize actions from 3D skeletons. The contextual updating mechanism of their LSTM units was further controlled by a gating mechanism which improved robustness. Núñez *et al.* [42] used a combination of convolutional neural networks (CNN) and LSTMs with a two-stage training process to classify skeleton and hand gestures. Avola *et al.* [3] used a LSTM architecture in conjunction with hand-crafted angular features of hand joints to recognize hand gestures.

In contrast, we only use gated recurrent units (GRU) [14] as the building block of our recurrent network. As we show later, GRUs are faster to train and produce better results. Also, our method is designed to be general and not specific to a particular device, gesture modality or feature representation. Lastly, we leverage the attention mechanism to capture the most important parts of each input sequence.

**Attention mechanism.** When using recurrent architectures, the sub-parts of a temporal sequence may not all be equally important: some subsequences may be more pertinent to the task at hand than others. Thus, it is often beneficial to learn a representation that can identify these important subsequences and leverage them to tackle the subject matter. This is the key intuition behind the attention model [4][40]. Even though the attention model was originally proposed for sequence to sequence models and neural machine translation, it has been adapted to the task of gesture and action recognition [1][6][7][22][38][51].

Liu *et al.* [38] proposed a global context-aware attention LSTM network for 3D action recognition. Using a global context, their method selectively focuses on the most informative joints when performing recognition. Song *et al.* [51] used the attention mechanism with LSTM units to selectively focus on discriminative skeleton joints at each gesture frame. Fan *et al.* [22] introduced a multiview re-observation LSTM network which augments any observed action with multiple views of the same action in order to achieve view-invariant recognition. Baradel *et al.* [6] proposed a two-stream convolutional and LSTM network which used pose as well as image information to perform action recognition. They demonstrated the importance of focusing on the hand motion of the actors in the sequence to improve recognition accuracy. Later, Baradel *et al.* [7] leveraged the visual attention model to recognize human activities purely using image data. They used GRUs as the building block of their recurrent architecture.

Contrary to some of this work, DeepGRU only requires pose and vector-based data. Our novel attention model differs from prior work in how the context vector is computed and consumed. For instance, GCA-LSTM [38] has a multi-pass attention subnetwork which requires multiple initialize/refine iterations to compute attention vectors. Ours is single-pass and not iterative. Our attention model also differs from STA-LSTM [51] which has two separate temporal and spatial components, whereas ours has only one component for both domains. VA-LSTM [71] has a view-adaptation subnetwork that learns transformations to consistent view-points. This imposes the assumption that input data are spatial or view-point dependent, which may prohibit applications on non-spatial data (*e.g.* acoustic gestures [45]). Our model does not make any such assumptions. As we show later, our single-pass, non-iterative, spatio-temporal combined attention, and device-agnostic architecture result in less complexity, fewer parameters, and shorter training time, while achieving state-of-the-art results, which we believe sets us apart from prior work.

## 3. DeepGRU

In this section we provide an in-depth discussion of DeepGRU's architecture. In our architecture, we take inspiration from VGG-16 [48], and the attention [4][40] and sequence to sequence models [52]. Our model, depicted in Figure 1, is comprised of three main components: an encoder network, the attention module, and two fully-connected (FC) layers fed to softmax producing the probability distribution of the class labels. We provide an ablation study to give insight into our design choices in Section 5.

### 3.1. Input Data

The input to DeepGRU is raw input device samples represented as a temporal sequence of the underlying gesture data (*e.g.* 3D joint positions, accelerometer or velocity measurements, 2D Cartesian coordinates of pen/touch interactions, *etc.*). At time step $t$, the input data is the column vector $x_t \in \mathbb{R}^N$, where $N$ is the dimensionality of the feature vector. Thus, the input data of the entire temporal sequence of a single gesture sample is the matrix $\mathbf{x} \in \mathbb{R}^{N \times L}$, where $L$ is the length of the sequence in time steps.

The dimensionality $N$ depends on the device that generated the data and also how one chooses to represent the data. In this sense, DeepGRU is agnostic to the input representation. For instance, consider a gesture sample collected from a Kinect device. This gesture sample might have the 3D position of 21 joints of a human actor's skeleton performing an action in $L$ time steps. One can take $N$ to be 3×21=63 dimensional and represent this sample as $\mathbf{x} \in \mathbb{R}^{63 \times L}$. Now consider a variation of this gesture sample that involves two human actors. In this case, one can take $N$ to be 2×3×21=126 dimensional (the sample as $\mathbf{x} \in \mathbb{R}^{126 \times L}$). Alternatively, one may choose to interleave the human skeletons temporally[1]. In this case, the dimensionality of $N$ would still be 63, however, the gesture sample itself would have double the number of time steps, making the sample $\mathbf{x} \in \mathbb{R}^{63 \times 2L}$.

Note that various input example sequences could have different number of time steps. We use the entire temporal sequence as-is without subsampling or clipping. When training on mini-batches, we represent the $i^{th}$ mini-batch as the tensor $\mathbf{X}_i \in \mathbb{R}^{B \times N \times \widetilde{L}}$, where $B$ is the mini-batch size and $\widetilde{L}$ is the length of the longest sequence in the $i^{th}$ mini-batch. Sequences that are shorter than $\widetilde{L}$ are zero-padded.

### 3.2. Encoder Network

The encoder network in DeepGRU is fed with data from training samples and serves as the feature extractor. Our encoder network consists of a total of five stacked unidirectional GRUs. Although LSTM units [25] are more prevalent in the literature, we utilize GRUs because due to the smaller number of parameters, these units are simpler to use and are generally faster to train and are less prone to overfitting. At time step $t$, given an input vector $x_t$ and the hidden state vector of the previous time step $h_{(t-1)}$, a GRU computes $h_t$, the hidden output at time step $t$, as $h_t = \Gamma\big(x_t, h_{(t-1)}\big)$ using the following transition equations:

$$r_t = \sigma\Big(\big(W_x^r\, x_t + b_x^r\big) + \big(W_h^r\, h_{(t-1)} + b_h^r\big)\Big) \quad (1)$$
$$u_t = \sigma\Big(\big(W_x^u\, x_t + b_x^u\big) + \big(W_h^u\, h_{(t-1)} + b_h^u\big)\Big)$$
$$c_t = \tanh\Big(\big(W_x^c\, x_t + b_x^c\big) + r_t\big(W_h^c\, h_{(t-1)} + b_h^c\big)\Big)$$
$$h_t = u_t \circ h_{(t-1)} + \Big(1 - u_t\Big) \circ c_t$$

where $\sigma$ is the sigmoid function, $\circ$ denotes the Hadamard product, $r_t$, $u_t$ and $c_t$ are reset, update and candidate gates respectively and $W_p^q$ and $b_p^q$ are the trainable weights and biases. In our encoder network, $h_0$ of all the GRUs are initialized to zero.

Given a gesture example $\mathbf{x} \in \mathbb{R}^{N \times L}$, the encoder network uses Equation 1 to output $\bar{h} \in \mathbb{R}^{128 \times L}$, where $\bar{h}$ is the result of the concatenation $\bar{h} = \big[h_0;\ h_1;\ ...;\ h_{(L-1)}\big]$. This output, which is a compact encoding of the input matrix $\mathbf{x}$, is then fed to the attention module.

---

[1] We chose to use this representation in our evaluations of multi-actor gestures.

### 3.3. Attention Module

The output of the encoder network, which is a compressed representation of the input gesture sample, can provide a reasonable set of features for performing classification. We further refine this set of features by extracting the most informative parts of the sequence using the attention model. We propose a novel adaptation of the global attention model [40] which is suitable for our recognition task.

Given all the hidden states $\bar{h}$ of the encoder network, our attention module computes the attentional context vector $c \in \mathbb{R}^{128}$ using the trainable parameters $W_c$ as:

$$c = \left( \frac{\exp\left(h_{(L-1)}^\mathsf{T} W_c \bar{h}\right)}{\sum_{t=0}^{L-1} \exp\left(h_{(L-1)}^\mathsf{T} W_c h_t\right)} \right) \bar{h} \qquad (2)$$

As evident in Equation 2, we solely use the hidden states of the encoder network to compute the attentional context vector. The hidden state of the last time step $h_{(L-1)}$ of the encoder network (the yellow arrow in Figure 1) is the main component of our context computation and attentional output. This is because $h_{(L-1)}$ can potentially capture a lot of information from the entire gesture sample sequence.

With the context vector at hand, one could use the concatenation $\bigl[c \,;\, h_{(L-1)}\bigr]$ to form the contextual feature vector and perform classification. However, recall that the inputs to DeepGRU can be of arbitrary lengths. Therefore, the amount of information that is captured by $h_{(L-1)}$ could differ among short sequences and long sequences. This could make the model susceptible to variations in sequence lengths. Our proposed solution to mitigate this is as follows. During training, we jointly learn a set of parameters that given the context and the hidden state of the encoder network would decide whether to use the hidden state directly, or have it undergo further transformation while accounting for the context. This decision logic can be mapped to the transition equations of a GRU (see Equation 1). Thus, after computing the context $c$, we additionally compute the auxiliary context $c'$ and produce the attention module's output $o_\text{attn}$ as follows:

$$\begin{aligned} c' &= \Gamma_\text{attn}(c, h_{(L-1)}) \\ o_\text{attn} &= \bigl[c \,;\, c'\bigr] \end{aligned} \qquad (3)$$

where $\Gamma_\text{attn}$ is the attentional GRU of the our model.

In summary, we believe that the novelty of our attention model is threefold. First, it only relies on the hidden state of the last time step $h_{(L-1)}$, which reduces complexity. Second, we compute the auxiliary context vector to mitigate the effects of sequence length variations. Lastly, our attention module is invariant to zero-padded sequences and thus can be trivially vectorized for training on mini-batches of sequences with different lengths. As we show in Section 5, our attention model works very well in practice.

### 3.4. Classification

The final layers of our model are comprised of two FC layers ($F_1$ and $F_2$) with ReLU activations that take the attention module's output and produce the probability distribution of the class labels using a softmax classifier:

$$\hat{y} = \text{softmax}\Bigl( F_2\Bigl(\text{ReLU}\bigl(F_1(o_\text{attn})\bigr)\Bigr)\Bigr) \qquad (4)$$

We use batch normalization [28] followed by dropout [24] on the input of both $F_1$ and $F_2$ in Equation 4. During training, we minimize the cross-entropy loss to reduce the difference between predicted class labels $\hat{y}$ and the ground truth labels $y$. More implementation details are discussed shortly.

## 4. Evaluation

To demonstrate the robustness and generality of Deep-GRU, we performed a set of experiments on datasets of various sizes. Specifically, we evaluate our proposed method on five datasets: UT-Kinect [69], NTU RGB+D [46], SYSU-3D [26], DHG 14/28 [15][17] and SBU Kinect Interactions [70]. We believe these datasets cover a wide range of gesture interactions, number of actors, view-point variations and input devices. We additionally performed experiments on two small-scale datasets (Wii Remote [11] and Acoustic [45]) in order to demonstrate the suitability of DeepGRU for scenarios where only a very limited amount of training data is available. We compute the recognition accuracies on each dataset and report them as a percentage.

**Implementation details.** We implemented DeepGRU using the PyTorch [44] framework. The input data to the network are z-score normalized using the training set. We use the Adam solver [32] ($\beta_1 = 0.9, \beta_2 = 0.999$) and the initial learning rate of $10^{-3}$ to train our model. The mini-batch size for all experiments is 128, except for those on NTU RGB+D, for which the size is 256. Training is done on a machine equipped with two NVIDIA GeForce GTX 1080 GPUs, Intel Core-i7 6850K processor and 32 GB RAM. Unless stated otherwise, both GPUs were used for training with mini-batches divided among both cards. We provide a reference implementation for the camera-ready version.

**Regularization.** We use dropout (0.5) and data augmentation to avoid overfitting. All regularization parameters were determined via cross-validation on a subset of the training data. Across all experiments we use three types of data augmentation: **(1)** random scaling with a factor[2] of $\pm 0.3$, **(2)** random translation with a factor of $\pm 1$, **(3)** synthetic sequence generation with gesture path stochastic resampling (GPSR) [55]. For GPSR we randomly select the resample count $n$ and remove count $r$. We use $n$ with a factor of $(\pm 0.1 \times \widetilde{L})$ and $r$ with a factor of $(\pm 0.05 \times \widetilde{L})$. Additionally, we use two more types of regularization for experiments on NTU RGB+D dataset. We use a weight decay value of $10^{-4}$, as well as random rotation with a factor of $\pm \frac{\pi}{4}$. This was necessary due to the multiview nature of the dataset.

| Method | Accuracy |
|---|---|
| Grassmann Manifold [49] | 88.5 |
| Histogram of 3D Joints [69] | 90.9 |
| Riemannian Manifold [18] | 91.5 |
| Key-Pose-Motifs [62] | 93.5 |
| LARP + mfPCA [2] | 94.8 |
| Action snippets [61] | 96.5 |
| ST LSTM + Trust Gates [37] | 97.0 |
| Lie Group [59] | 97.1 |
| Graph-based [65] | 97.4 |
| ST-NBNN [67] | 98.0 |
| SCK + DCK [33] | 98.2 |
| DPRL + GCNN [53] | 98.5 |
| GCA-LSTM *(direct)* [38] | 98.5 |
| CNN + Kernel Feature Maps [57] | 98.9 |
| GCA-LSTM *(stepwise)* [38] | 99.0 |
| CNN + LSTM [42] | 99.0 |
| KRP FS [13] | 99.0 |
| **DeepGRU** | **100.0** |

Table 1. Results on UT-Kinect [69] dataset.

### 4.1. UT-Kinect

This dataset [69] is comprised of ten gestures performed by ten participants two times (200 sequences in total). The data of each participant is recorded and labeled in one continuous session. What makes this dataset challenging is that the participants move around the scene and perform the gestures consecutively. Thus, samples have different starting position and/or orientations. We use the leave-one-out-sequence cross validation protocol of [69]. During our tests, we noticed that the label of one of the sequences was corrupted[3]. We manually labeled the sequence and performed our experiments twice: once with the corrupted sequence omitted, and once with our manually labeled version of the corrupted sequence. We obtained the same results in both settings. Our approach achieves state-of-the-art results with the perfect classification accuracy of 100% as shown in Table 1.

---

[2]A factor of $\pm 0.3$ indicates that samples are randomly and non-uniformly (*e.g.*) scaled along all axes to [0.7, 1.3] of their original size

[3]The second example of participant 10's *carry* gesture

| Modality | Method | Accuracy CS | Accuracy CV |
|---|---|---|---|
| Image | Multitask DL [41] | 84.6 | – |
|  | **Glimpse Clouds [7]** | **86.6** | **93.2** |
| Pose+Image | DSSCA - SSLM [47] | 74.9 | – |
|  | STA Model (Hands) [5] | 82.5 | 88.6 |
|  | Hands Attention [6] | 84.8 | 90.6 |
|  | **Multitask DL [41]** | **85.5** | – |
| Pose | Skeletal Quads [21] | 38.6 | 41.4 |
|  | Lie Group [59] | 50.1 | 52.8 |
|  | HBRNN [20] | 59.1 | 64.0 |
|  | Dynamic Skeletons [26] | 60.2 | 65.2 |
|  | Deep LSTM [46] | 60.7 | 67.3 |
|  | Part-aware LSTM [46] | 62.9 | 70.3 |
|  | ST LSTM + Trust Gates [37] | 69.2 | 77.7 |
|  | STA Model [51] | 73.2 | 81.2 |
|  | LSTM + FA + VF [22] | 73.8 | 85.9 |
|  | Temporal Sliding LSTM [36] | 74.6 | 81.3 |
|  | CNN + Kernel Feature Maps [57] | 75.3 | – |
|  | SkeletonNet [30] | 75.9 | 81.2 |
|  | GCA-LSTM *(direct)* [38] | 74.3 | 82.8 |
|  | GCA-LSTM *(stepwise)* [38] | 76.1 | 84.0 |
|  | JTM CNN [64] | 76.3 | 81.1 |
|  | DPTC [66] | 76.8 | 84.9 |
|  | VA-LSTM [71] | 79.4 | 87.6 |
|  | Beyond Joints [63] | 79.5 | 87.6 |
|  | Clips+CNN+MTLN [31] | 79.6 | 84.8 |
|  | View-invariant [39] | 80.0 | 87.2 |
|  | Dual Stream CNN [68] | 81.1 | 87.2 |
|  | DPRL + GCNN [53] | 83.5 | 89.8 |
|  | **DeepGRU** | **84.9** | **92.3** |

Table 2. Results on NTU RGB+D [46] dataset.

### 4.2. NTU RGB+D

To our knowledge, this is the largest dataset of actions collected from Kinect (v2) [46]. It comprises about 56,000 samples of 60 action classes performed by 40 subjects. Each subject's skeleton has 25 joints. The challenging aspect of this dataset stems from the availability of various viewpoints for each action, as well as the multi-person nature of some action classes. We follow the cross-subject (CS) and cross-view (CV) evaluation protocols of [46]. In the CS protocol, 20 subjects are used for training and the remaining 20 subjects are used for testing. In the CV protocol, two viewpoints are used for training and the remaining one viewpoint is used for testing. Note that according to the dataset authors, 302 samples in this dataset have missing or incomplete skeleton data which were omitted in our tests.

We create our feature vectors similar to [46]. For each action frame, we concatenate the 3D coordinates of the skeleton joints into one 75 dimensional vector in the order that they appear in the dataset. In cases where there are multiple skeletons in a single action frame, we treat each skeleton as one single time step. For each frame, we detect the main actor, which is the skeleton with the largest amount of total skeleton motion. The time step frames are created in descending order of total skeleton motion. Following [46], we transform the coordinates of all skeletons to the spine-mid joint of the main actor in the action frame.

| Method | Accuracy |
|---|---|
| LAAF [27] | 54.2 |
| Dynamic Skeletons [26] | 75.5 |
| ST LSTM + Trust Gates [37] | 76.5 |
| DPRL + GCNN [53] | 76.9 |
| VA-LSTM [71] | 77.5 |
| GCA-LSTM *(direct)* [38] | 77.8 |
| GCA-LSTM *(stepwise)* [38] | 78.6 |
| **DeepGRU** | **80.3** |

Table 3. Results on SYSU-3D [26].

| Protocol | Method | Accuracy | |
|---|---|---|---|
| | | C = 14 | C = 28 |
| Leave-one-out | Chen *et al.*[12] | 84.6 | 80.3 |
| | De Smedt *et al.*[16] | 82.5 | 68.1 |
| | CNN+LSTM [42] | 85.6 | 81.1 |
| | DPTC [66] | 85.8 | 80.2 |
| | **DeepGRU** | **92.0** | **87.8** |
| SHREC'17 [17] | HOG$^2$ [43][17] | 78.5 | 74.0 |
| | HIF3D [9] | 90.4 | 80.4 |
| | De Smedt *et al.*[50][17] | 88.2 | 81.9 |
| | Devineau *et al.*[19] | 91.2 | 84.3 |
| | **DLSTM** [3] | **97.6** | **91.4** |
| | **DeepGRU** | 94.5 | 91.4 |

Table 4. Results on DHG 14/28 [15] with two evaluation protocols.

| Modality | Method | Accuracy |
|---|---|---|
| Image | Hands Attention [6] | 72.0 |
| | DSPM | 93.4 |
| Pose + Image | Hands Attention [6] | 94.1 |
| Pose | HBRNN [20] | 80.4 |
| | Deep LSTM [46] | 86.0 |
| | Ji *et al.*[29] | 86.8 |
| | Co-occurance Deep LSTM [72] | 90.4 |
| | Hands Attention [6] | 90.5 |
| | STA Model [51] | 91.5 |
| | ST LSTM + Trust Gates [37] | 93.3 |
| | SkeletonNet [30] | 93.5 |
| | Clips + CNN + MTLN [31] | 93.5 |
| | GCA-LSTM *(direct)* [38] | 94.1 |
| | CNN + Kernel Feature Maps [57] | 94.3 |
| | GCA-LSTM *(stepwise)* [38] | 94.9 |
| | LSTM + FA + VF [22] | 95.0 |
| | **VA-LSTM** [71] | **97.2** |
| | DeepGRU | 95.7 |

Table 5. Results on SBU Kinect Interactions [70].

Our results are presented in Table 2. Although DeepGRU only uses the raw skeleton positions of the samples, we present the results of other recognition methods that use other types of gesture data. To the best of our knowledge, DeepGRU achieves state-of-the-art performance among all methods that only use raw skeleton pose data.

### 4.3. SYSU-3D

This Kienct-based dataset [26] contains 12 gestures performed by 40 participants totaling 480 samples. The widely-adopted evaluation protocol [26] of this dataset is to randomly select 20 subjects for training and the use remaining 20 subjects for testing. This process is repeated 30 times and the results are averaged. The results of our experiments are presented in Table 3.

### 4.4. DHG 14/28

This dataset [15] contains 14 hand gestures of 28 participants collected by a near-view Intel RealSense depth camera. Each gesture is performed in two different ways: using the whole hand, or just one finger. Also, each example gesture is repeated between one to ten times yielding 2800 sequences. The training and testing data on this dataset are predefined and evaluation can be performed in two ways: classify 14 gestures or classify 28 gestures. The former is insensitive to how an action is performed, while the latter discriminates the examples performed with one finger from the ones performed with the whole hand. The standard evaluation protocol of this dataset is a leave-one-out cross-validation protocol. However, SHREC 2017 [17] challenge introduces a secondary protocol in which training and testing sets are pre-split. Table 4 depicts our results using both protocols and both number of gesture classes.

### 4.5. SBU Kinect Interactions

This dataset [70] contains 8 two-person interactions of seven participants. We utilize the 5-fold cross-validation protocol of [70] in our experiments. Contrary to other datasets, which express joint coordinates in the world coordinate system, this dataset has opted to normalize the joint values instead. Despite using a Kinect (v1) sensor, the participants in the dataset have only 15 joints.

We treat action frames that contain multiple skeletons similarly to what we described above for the NTU RGB+D dataset, with the exception of transforming the joint coordinates. Also, using the equations provided in the datasets, we covert the joint values them to metric coordinates in the depth camera coordinate frame. This is necessary to make the representation consistent with other datasets that we experiment on. Table 5 summarizes our results.

## 4.6. Small Training Set Evaluation

The amount of training data for some gesture-based applications may be limited. This is especially the case during application prototyping stages, where developers tend to rapidly iterate through design and evaluation cycles. Throughout the years, various methods have been proposed in the literature aiming to specifically address the need for recognizers that are easy to implement, fast to train and work well with small training sets [34] [35] [54] [56].

Traditionally, deep networks are believed to be slow to train, requiring a lot of training data. We show this is not the case with DeepGRU and our model performs well with small training sets and can be trained only on the CPU. We pit DeepGRU against Protractor3D [35], $3 [34] and Jackknife [56] which to our knowledge produce high recognition accuracies with a small number of training examples [56].

We examine two datasets. The first dataset contains acoustic over-the-air hand gestures via Doppler shifted soundwaves [45]. This dataset contains 18 hand gestures collected from 22 participants via five speakers and one microphone. At 165 component vectors per frame, this dataset is very high-dimensional. Also, the soundwave-based interaction modality is prone to high amounts of noise. The second dataset contains gestures performed via a Wii Remote controller [11]. This dataset contains 15625 gestures of 25 gesture classes collected from 25 participants. In terms of data representation, both datasets differ from all others examined thus far. Samples of [45] are frequency binned spectrograms while samples of [11] are linear acceleration data and angular velocity readings (6D), neither of which resemble typical skeletal representations nor positional features.

For each experiment we use the user-dependent protocol of [11][56]. Given a particular participant, $\mathcal{T}$ random samples from that participant are selected for training and the remaining samples are selected for testing. This procedure is repeated per participant and the results are averaged across all of them. Considering that in the prototyping stage the amount of training samples is typically limited, we evaluate the performance of all the recognizers using $\mathcal{T}=2$ and $\mathcal{T}=4$ training samples per gesture class. These results are tabulated in Table 6. Even though deep networks are not commonly used with very small training sets, DeepGRU demonstrates very competitive accuracy in these tests. We see that with $\mathcal{T}=4$ training samples per gesture class, DeepGRU outperforms other recognizers on both datasets.

## 5. Discussion

**Comparison with the state-of-the-art.** Experiment results show that DeepGRU generally tends to outperform the state-of-the-art results, sometimes with a large margin. On the NTU-RGB+D [46], we observe that in some cases DeepGRU outperforms image-based or hybrid methods.

| Dataset | Method | Accuracy | |
|---|---|---|---|
| | | $\mathcal{T}=2$ | $\mathcal{T}=4$ |
| Acoustic [45] | **Jackknife** [56] | **91.0** | 94.0 |
| | **DeepGRU** | 89.0 | **97.4** |
| Wii Remote [11] | Protractor3D [35] | 73.0 | 79.6 |
| | $3 [34] | 79.0 | 86.1 |
| | **Jackknife** [56] | **96.0** | 98.0 |
| | **DeepGRU** | 92.4 | **98.3** |

Table 6. Rapid prototyping evaluation results with $\mathcal{T}$ training samples per gesture class.

| Device | Configuration | Dataset | Time (mins) |
|---|---|---|---|
| CPU | 12 threads | Acoustic [45] ($\mathcal{T}=4$) | 1.7 |
| | | Wii Remote [11] ($\mathcal{T}=4$) | 6.9 |
| GPU | 2× GTX 1080 | SHREC 2017 [17] | 5.5 |
| | | NTU RGB+D [46] | 129.6 |
| | 1× GTX 1080 | SHREC 2017 [17] | 6.2 |
| | | SYSU-3D [26] | 9.0 |
| | | NTU RGB+D [46] | 198.5 |

Table 7. DeepGRU training times (in minutes) on various datasets.

Although the same superiority is observed on the SBU dataset [70], our method achieves slightly lower accuracy compared to VA-LSTM [71]. One possible intuition for this observation could be that the SBU dataset [70] provides only a subset of skeleton joints that a Kinect (v1) device can produce (15 compared to the full set of 20 joints). Further, note that VA-LSTM's view-adaptation subnetwork assumes that the gesture data are 3D positions and viewpoint-dependent. This is in contrast with DeepGRU which does not make such assumptions about the underlying type of the input data (position, acceleration, velocity, *etc.*).

As shown in Table 4, classifying 14 gestures of the DHG 14/28 dataset [15] with DLSTM [3] yields higher recognition accuracy compared to DeepGRU. As previously mentioned, DLSTM [3] uses hand-crafted angular features extracted from hand joints and these features are used as the input to the recurrent network while DeepGRU uses raw input, which relieves the user of the burden of computing domain-specific features. Classifying 28 classes, however, yields similar results with either of the recognizers.

**Generality.** Our experiments demonstrate the versatility of DeepGRU for various gesture or action modalities and input data: from full-body multi-actor actions to hand gestures, collected from various commodity hardware such as depth sensors or game controllers with various data representations (*e.g.* pose, acceleration and velocity or frequency spectrograms) as well as other differences such as the number of actors, gesture lengths, number of samples and number of viewpoints. Regardless of these differences, DeepGRU can still produce high accuracy results.

This flexibility is, in large part, due to our attention mod-ule and context vector computation. We present an example visualization of our attention module's response in Figure 2. We can see that after training, our attention mechanism correctly selects the most discriminative frames in the sequence.

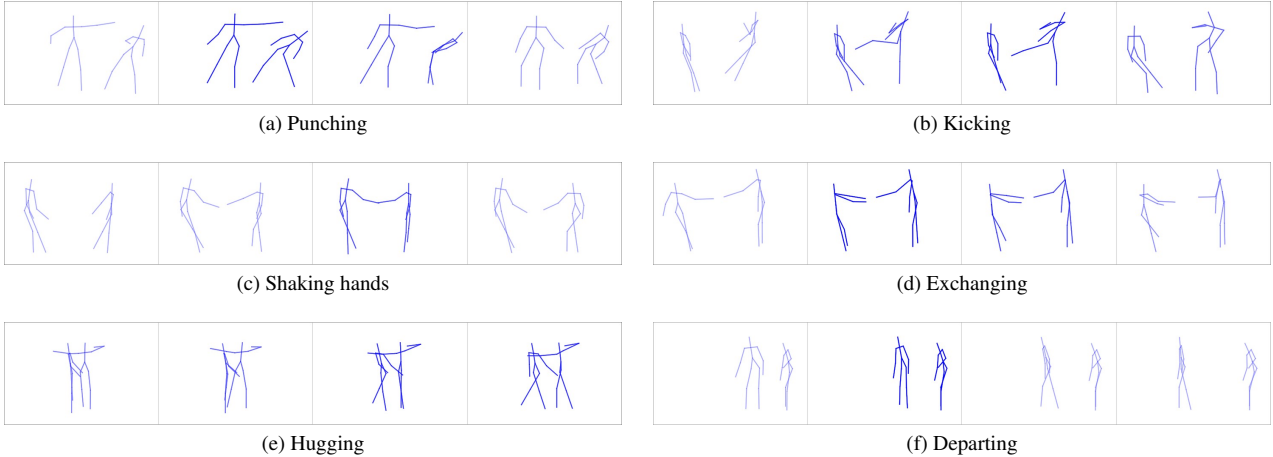

Figure 2. Attention response visualization for samples from the SBU Kinect Interactions [70] dataset. Each sample starts from the left and progresses to the right. The color intensity indicates the amount of attentional response (norm) to the frame (darker = higher response).

**Ease of use.** In addition to accuracy, the adoption of any one gesture recognition method ultimately comes down to the ease of use. In that regards, DeepGRU has a few advantages over competitive methods. Our method uses raw device data, thus requiring fairly little domain knowledge. Our model is straightforward to implement and as we discuss shortly, training is fast. We believe these traits make DeepGRU an enticing option for practitioners.

**Timings.** Training times is an important factor in the prototyping stage. In such scenarios, the ability to conveniently train a network without GPUs is desirable. We measured the amount of time it takes to train DeepGRU to convergence with different configurations in Table 7. The reported times include dataset loading, preprocessing and data augmentation time. Training our model to convergence tends to be fast. In fact, GPU training of medium-sized datasets or CPU-only training of small datasets can be done in under 10 minutes, which we believe is beneficial for iterative design. We also measured DeepGRU's average inference time per sample both on GPU and on CPU in *micro*seconds. On a single GPU, our methods takes 349.1 µs to classify one gesture example while it takes 3136.3 µs on the CPU.

**Ablation study.** To provide insight into our network design, we present an ablation study in Table 8. Most importantly, we note depth alone is not sufficient to achieve state-of-the-art results. Further, accuracy increases in all cases when we use GRUs instead of LSTMs. GRUs were on average 12% faster to train and the worst GRU variant achieved higher accuracy than the best LSTM one. In our early experiments we noted LSTM networks overfitted frequently which necessitated a lot more parameter tuning, motivating our preference for GRUs. However, we later observed underfitting when training GRU variants on larger datasets, arising the need to reduce regularization and tune parameters again. To alleviate this, we added the second FC layer which later showed to improve results across all datasets while still faster than LSTMs to train. We observe increased accuracy in all experiments with attention, which suggests the attention model is necessary. Lastly, in our experiments we observed an improvement of roughly 0.5%–1% when the auxiliary context vector is used (Section 3.3). In short, we see improved results with the attention model on GRU variants with five stacked layers and two FC layers.

**Limitations.** Our method has some limitations. Most importantly, the input needs to be segmented, although adding support for unsegmented data is straightforward, requiring a change in the training protocol as demonstrated in [10]. In our experiments we observed that DeepGRU typically performs better with high-dimensional data, thus application on low-dimensional data may require further effort from developers. Although we used a similar set of hyperparameters for all experiments, other datasets may require some tuning.

## 6. Conclusion

We discussed DeepGRU, a deep network-based gesture and action recognizer which directly works with raw pose and vector data. We demonstrated that our architecture, which uses stacked GRU units and a global attention mechanism along with two fully-connected layers, was able to achieve state-of-the-art recognition results on various datasets, regardless of the dataset size and interaction modality. We further examined our approach for application in scenarios where training data is limited and computational power is constrained. Our results indicate that with as little as four training samples per gesture class, DeepGRU can still achieve competitive accuracy. We also showed that training times are short and CPU-only training is possible.

| Attn. | Rec. Unit | # Stacked | # FC | Time (sec) | Accuracy | Attn. | Rec. Unit | # Stacked | # FC | Time (sec) | Accuracy |
|---|---|---|---|---|---|---|---|---|---|---|---|
| - | LSTM | 3 | 1 | 162.21 | 91.78 | ✓ | LSTM | 3 | 1 | 188.29 | 92.74 |
| - | LSTM | 3 | 2 | 164.07 | 91.07 | ✓ | LSTM | 3 | 2 | 192.12 | 92.02 |
| - | LSTM | 5 | 1 | 246.47 | 91.90 | ✓ | LSTM | 5 | 1 | 277.32 | 92.38 |
| - | LSTM | 5 | 2 | 251.67 | 89.52 | ✓ | LSTM | 5 | 2 | 283.35 | 92.26 |
| - | GRU | 3 | 1 | 143.87 | 93.45 | ✓ | GRU | 3 | 1 | 170.48 | 94.12 |
| - | GRU | 3 | 2 | 148.08 | 93.33 | ✓ | GRU | 3 | 2 | 174.00 | 93.81 |
| - | GRU | 5 | 1 | 210.83 | 93.69 | ✓ | GRU | 5 | 1 | 243.10 | 93.93 |
| - | GRU | 5 | 2 | 212.99 | 93.81 | ✓ | **GRU** | **5** | **2** | **248.66** | **94.52** |

Table 8. Ablation study on DHG 14/28 dataset (14 class, SHREC'17 protocol). We examine (respectively) the effects of the usage of the attention model, the recurrent layer choice (LSTM *vs.* GRU), the number of stacked recurrent layers (3 *vs.* 5) and the number of FC layers (1 *vs.* 2). Training times (seconds) are reported for every model. Experiments use the same random seed. DeepGRU's model is boldfaced.

As for future direction, we plan to extend our method to support other types of data, such as images and videos. The availability of additional data would likely increase the robustness of DeepGRU. We also intend to extend our method to support unsegmented data streams, which should broaden the range of application scenarios for our method. Finally, a detailed study of the effects of data dimensionality as well as feature representation on the performance of DeepGRU would aid focusing on what may need further improvement.

## Acknowledgement

We thank Eugene M. Taranta II, Amir Gholaminejad, Alaleh Razmjoo, Mehdi Sajjadi and Kihwan Kim for the helpful discussions and the ICE lab members at UCF for their support. Portions of this research used the NTU RGB+D Action Recognition Dataset [46] made available by the ROSE Lab at the Nanyang Technological University, Singapore.